\documentclass{ifacconf}
\usepackage{amsfonts}
\usepackage{amsmath,bm}
\usepackage{graphicx}      % include this line if your document contains figures
\graphicspath{ {./figures/} }
\usepackage{natbib}        % required for bibliography
\begin{document}
\begin{frontmatter}

\title{Learning Time Delay Systems with Neural Ordinary Differential Equations} 

\author[*]{Xunbi~A.~Ji}
and
\author[*,**]{G\'{a}bor~Orosz}

\address[*]{Department of Mechanical Engineering, 
   }
\address[**]{Department of Civil and Environmental Engineering, \\
University of Michigan, Ann Arbor, MI 48109 USA}
\address{(e-mail: xunbij@umich.edu, orosz@umich.edu)}

\begin{abstract}                
% Abstract of not more than 250 words.
A novel way of using neural networks to learn the dynamics of time delay systems from sequential data is proposed. A neural network with trainable delays is used to approximate the right hand side of a delay differential equation. We relate the delay differential equation to an ordinary differential equation by discretizing the time history and train the corresponding neural ordinary differential equation (NODE) to learn the dynamics. 
An example on learning the dynamics of the Mackey-Glass equation using data from chaotic behavior is given. After learning both the nonlinearity and the time delay, we demonstrate that the bifurcation diagram of the neural network matches that of the original system. 
\end{abstract}

\begin{keyword}
neural ordinary differential equations, time delay systems, time delay neural networks, trainable time delays, Mackey-Glass equation.
\end{keyword}

\end{frontmatter}
%===============================================================================

%%%%%%%%%%%%%%%%%%%%%%%%%%%%%%%%%%%%%%%%%%%%%%%%%%%%%%%%%%%%
\section{Introduction}
%%%%%%%%%%%%%%%%%%%%%%%%%%%%%%%%%%%%%%%%%%%%%%%%%%%%%%%%%%%%
\vspace{-3mm}

Artificial neural networks have been thriving in many fields over the past decade and their power in approximating and generalizing the underlying input-output relationships has been demonstrated for multiple examples. 
Multiple research studies related neural networks to dynamical systems; see, for instance, \cite{kumpati1990identification,pei2013mapping,brunton2019data}.

Knowledge from dynamical systems can help to design and train neural networks and many ideas in recurrent neural networks were inspired by dynamical systems principles. 
%Networks can represent discrete time or continuous time dynamics depending on whether they describe the relationship between sequential states or the relationship between the states and their time derivatives. 
The deep equilibrium model (DEM) in~\cite{bai2019deep} represents the discrete-time map between hidden states. Since a globally stable autonomous dynamical system approaches a fix point, one can directly solve for the fix point via root-finding.
Neural ordinary differential equation (NODE) in~\cite{chen2018neural} expresses the time derivative (i.e., the right hand side of the differential equation), while the evolution of the network for a given time period provides the map from the initial condition (input) and the terminal state (target).
The depth of the network is related to the simulation time of the corresponding ordinary differential equation.
Neural networks can be used to learn the dynamical systems. 
For example, \cite{Wong2021} provides an example of using discrete-time recurrent neural network to learn traffic dynamics while \cite{li2005approximation} uses a continuous-time recurrent neural network to express the right hand side of a differential equation.
Also, see \cite{turan2021multiple,rahman2022neural,koch2022physics} for some recent applications of NODEs on learning nonlinear dynamical systems.

In case of delayed dynamical systems, one shall include time delays in the neural networks. As a matter of fact, incorporating time delays in the neural networks may increase the capabilities of the networks. 
For example, in the neural delay differential equation (NDDE) approach proposed by~\cite{zhu2021neural}, adding the delayed state, the network is able to approximate trajectories that intersect in the original state space. One may increase the dimension of the system by defining extra states and construct an augmented ordinary differential equation (ANODE), as in~\cite{dupont2019augmented}, but this may not be able to capture the physical meaning of time delays in data-based systems.

Although time delays have been introduced to neural networks in both discrete time~(\cite{waibel1989phoneme,zhang2022improved}) and continuous time~(\cite{marcus1989stability,stelzer2021deep}), these delays were fixed rather than treated as parameters to be learned.
In this work, we consider a network with trainable time delays, a concept that was originally proposed in~\cite{ji2020}. 
Different from NDDE, we use an ordinary differential equation
to approximate the delayed dynamics, and thus, methods developed for training NODE can be directly applied.
This construction is based on the idea of approximating the infinitesimal generator through discretizing the history as in~\cite{breda2014stability}. 
For data-based models, our method bridges the gap between infinite-dimensional time delay systems and discrete-time maps obtained via full discretization.

The layout of the paper is as follows.
In Section~\ref{sec:discretization}, we relate delay differential equations (DDEs) to ordinary differential equations (ODEs) via discretization of the history. In Section~\ref{sec:TDNN}, we introduce trainable time delay neural networks and construct the NODE by discretizing the history. In Section~\ref{sec:train} we illustrate the training algorithm with a specific loss function. We provide an example of learning the Mackey-Glass equation from simulation data in Section~\ref{sec:MG} and conclude the results in  Section~\ref{sec:conclu}.

%%%%%%%%%%%%%%%%%%%%%%%%%%%%%%%%%%%%%%%%%%%%%%%%%%%%%%%%%%%%
\section{Discritization of the history} \label{sec:discretization}
%%%%%%%%%%%%%%%%%%%%%%%%%%%%%%%%%%%%%%%%%%%%%%%%%%%%%%%%%%%%

Consider an autonomous time delay system 
\begin{equation}\label{eq:DDE}
    \dot{x}(t) = g(x_t),
\end{equation}
where ${x\in \mathbb{R}^{n}}$ and ${x_t \in \mathcal{X}}$. Here ${\mathcal{X} = C([-\tau_{\rm max},0],\mathbb{R}^{n})}$ is the space of continuous functions, i.e., ${x_t(\vartheta) := x(t+\vartheta)}$, ${\vartheta\in[-\tau_{\max},0]}$, and ${ \tau_{\max}>0}$ is the maximum delay. The system \eqref{eq:DDE} contains the functional ${g \colon \mathcal{X} \to \mathbb{R}^{n} }$ and it can be re-written as an operator differential equation:
\begin{equation}\label{eq:AODE}
    \dot{x}_t = \mathcal{G}(x_t),
\end{equation}
where the operator ${\mathcal{G}: \mathcal{X}\rightarrow \mathcal{X}}$ is defined as
\begin{equation}\label{eq:AODE2}
    \mathcal{G}(\varphi) = \begin{cases}
                g(\varphi), \quad & \vartheta = 0,
               \\ \frac{\d \varphi}{\d \vartheta}, & \vartheta\in [-\tau_{\rm max},0),
                \end{cases}
\end{equation}
see \cite{stepan1989,breda2014stability}.

We discretize in the variable $\vartheta$ using a mesh of size ${M+1}$ and define the vector 
\begin{equation}\label{eq:bigX}
   X(t) = 
      \begin{bmatrix} 
   X_1(t) \\ X_2(t) \\ \vdots \\ X_{M+1}(t)
   \end{bmatrix} 
   =
   \begin{bmatrix} 
   x(t) \\ x(t-h) \\ \vdots \\ x(t-Mh)
   \end{bmatrix} 
\end{equation}
where ${h = \tau_{\rm max}/M}$. This way we can approximate \eqref{eq:AODE} with the ordinary differential equation
\begin{equation}\label{eq:AgmODE}
    \dot{X}(t) = G\big(X(t)\big),
\end{equation}
where the function ${G\colon \mathbb{R}^{n(M+1)} \rightarrow \mathbb{R}^{n(M+1)}}$ is given by
\begin{equation}\label{eq:AgmODE2}
    G(X) = \begin{bmatrix} \tilde{g}(X)\\
    D_M X\end{bmatrix}.
\end{equation}

Here ${\tilde{g} \colon \mathbb{R}^{n(M+1)} \rightarrow \mathbb{R}^{n}}$ is the approximation of the functional ${g \colon \mathcal{X} \to \mathbb{R}^{n} }$. For example, considering 
\begin{equation}\label{eq:example}
g(x_t) = f\big(x(t), x(t-(j+\alpha)h), x(t-\tau_{\rm max})\big)
\end{equation}
with $j\in\{0,1,\ldots,M-1\} $, $\alpha\in (0,1)$, 
and ${f\colon \mathbb{R}^{3n} \to \mathbb{R}^{n}}$, one may use linear interpolation to obtain 
\begin{equation}\label{eq:example2}
x(t-(j+\alpha)h)=(1-\alpha)x(t-jh)+\alpha x(t-(j+1)h),
\end{equation}
yielding
\begin{equation}\label{eq:example3}
\tilde{g}(X) = f\big(X_1, (1-\alpha)X_{j+1} + \alpha X_{j+2}, X_{M+1} \big),
\end{equation} 

The matrix ${D_M \in \mathbb{R}^{(n-1)(M+1)\times n(M+1)}}$ represents the numerical differentiation scheme used to approximate the time derivative of past states. For instance, choosing the forward Euler method to approximate the time derivative 
\begin{equation}
\dot{x}(t-jh) = \textstyle \frac{1}{h}\big(x(t-(j-1)h)-x(t-jh)\big),
\end{equation}
gives
\begin{equation}
\dot{X}_{j+1}(t) = \textstyle \frac{1}{h}\big(X_j(t)-X_{j+1}(t)\big),
\end{equation}
yielding
\begin{equation}\label{eq:DM}
D_M = \frac{1}{h}
\begin{bmatrix}
I & -I &        &        &  \\
  &  I & -I     &        &  \\
  &    & \ddots & \ddots &  \\
  &    &        & I      & -I 
\end{bmatrix},
\end{equation}
where $I \in \mathbb{R}^{n \times n}$ is the $n$-dimensional unit matrix and only the nonzero blocks of $D_M$ are spelled out.

\begin{figure}
\begin{center}
\includegraphics[width=8cm]{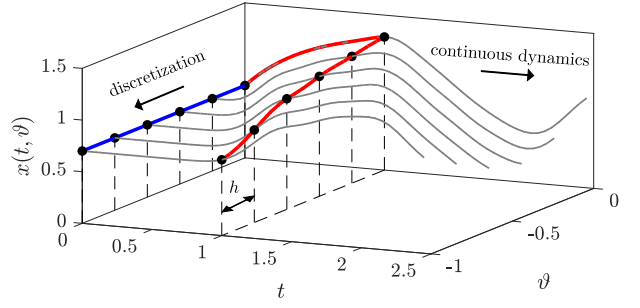}
\caption{Illustration of discretization of a DDE in history.} 
\label{fig:pde_sketch}
\end{center}
\end{figure}

In summary, (\ref{eq:AgmODE},\ref{eq:AgmODE2}) discretizes (\ref{eq:AODE},\ref{eq:AODE2}) in the history variable $\vartheta$. As illustrated in Fig.~\ref{fig:pde_sketch}, we still maintain the continuous dynamics in time $t$. This enables us to propagate the state forward with high accuracy, which will be beneficial for learning. Note that, however, data samples will only be available with sampling time $h$. 

%%%%%%%%%%%%%%%%%%%%%%%%%%%%%%%%%%%%%%%%%%%%%%%%%%%%%%%%%%%%
\section{Time delay neural networks}
\label{sec:TDNN}
%%%%%%%%%%%%%%%%%%%%%%%%%%%%%%%%%%%%%%%%%%%%%%%%%%%%%%%%%%%%

Time delay neural networks (TDNNs) were first proposed for speech recognition in~\cite{waibel1989phoneme}. In particular, networks with input delays can be expressed by
\begin{equation}\label{eq:TDNN}
    \begin{split}
    z^t_{1}&=f(W_{1}\mathbf{z}^t_{0}+b_{1}),\\
        z^t_l &= f_l\big(W_l z^t_{l-1} + b_l\big), \quad l \geq 2,
    \end{split}
\end{equation}
where ${l=1,2,\ldots,L}$ denotes the layers of the neural network while $t$ denotes time. The inputs for the network are the delayed values of the signal $z_0^t$ which can be collected in the vector
\begin{equation}\label{eq:TDNN2}
    \mathbf{z}^t_{0} = 
\begin{bmatrix}
z^{t-\tau_1}_{0} \\ 
\vdots \\
z^{t-\tau_d}_{0}
\end{bmatrix},
\end{equation} with input delays ${0\leq\tau_1<\cdots<\tau_d}$. The dimension of ${z_l}$ depends on the number of neurons in each layer, and the network output $z_L$ has the same dimension as $z_0$. The matrix $W_l$ and the vector $b_l$ are referred to as weight and bias respectively, while $f_l$ denotes the element-wise nonlinear function in each layer. Building on our previous work in \cite{ji2021}, we use trainable time delay neural networks approximate the right hand side of the delay differential equations. 
We shape the network indirectly by propagating the states of the approximate system forward via numerical simulation (using time steps much smaller than $h$) and comparing the result to data (which is available with sampling time $h$). This provides an efficient way of learning the nonlinear dynamics and the delays. 

In order to use the construction (\ref{eq:TDNN},\ref{eq:TDNN2}) for learning the dynamics (\ref{eq:AgmODE},\ref{eq:AgmODE2}), we define ${z_0^t = x(t)}$. We assume that the number of delays $d$ in the system is known (or at least the upper limit of this number is known). Note that, however, the values of the delays ${0\leq\tau_1<\cdots<\tau_d}$ are not known. Then we define a matrix which allows us to approximate the delayed states $z_0^{t-
\tau_i}$ in \eqref{eq:TDNN2} with the components $X_j$ in \eqref{eq:bigX}, namely
\begin{equation}\label{eq:P}
\mathbf{z}^t_{0} = P X(t),    
\end{equation}
where ${P\in\mathbb{R}^{d \times n(M+1)}}$ and since typically $d \ll M$, $P$ is a wide matrix.
For the case of the linear interpolation given in example (\ref{eq:example},\ref{eq:example2},\ref{eq:example3}) we have
\begin{equation}\label{eq:Pex}
P=
\begin{bmatrix}
I & \qquad &             &           &               & \\
  &        & (1-\alpha)I &  \alpha I &               & \\
  &        &             &          & \qquad \qquad & I
\end{bmatrix}
\end{equation}
where the block $(1-\alpha)I$ is the $j+1$-st column and the block $\alpha I$ is the $j+2$-nd column. Indeed, one may use higher-order interpolation methods which result in more nonzero blocks in $P$. 

The output of the neural network \eqref{eq:TDNN} is set to be the derivative
\begin{equation}\label{eq:network}
\dot{x}(t) = z^t_L = {\rm net}\big(P X(t)\big),    
\end{equation}
and thus the neural ordinary differential equation becomes
\begin{equation}\label{eq:NODE}
    \dot{X}(t) = \widehat{G}\big(X(t)\big),
\end{equation}
where the function ${\widehat{G}\colon \mathbb{R}^{n(M+1)} \rightarrow \mathbb{R}^{n(M+1)}}$ is given by
\begin{equation}\label{eq:NODE2}
    \widehat{G}(X) = 
    \begin{bmatrix} 
    {\rm net}\big(PX\big) \\
    D_M X
    \end{bmatrix},
\end{equation}
cf.~(\ref{eq:AgmODE},\ref{eq:AgmODE2}). We remark that to obtain $D_M$, instead of the forward Euler method we will use the central difference method which yields
\begin{equation}\label{eq:DM2}
D_M = \frac{1}{2h}
\begin{bmatrix}
I & O       & -I      &        &       \\
  &  \ddots &  \ddots & \ddots &       \\
  &         & I       & O      & -I    \\
  &         &         & 2I     & -2I  
\end{bmatrix},
\end{equation}
where $O \in \mathbb{R}^{n \times n}$ is the $n$-dimensional zero matrix while the other nonzero blocks are not spelled out; cf.~\eqref{eq:DM}. Note that the last row in $D_M$ is still derived from forward Euler method.

When training the NODE (\ref{eq:NODE},\ref{eq:NODE2}), the weights $W_l$ and biases $b_l$ in \eqref{eq:TDNN} as well as the delays $\tau_i$ are updated at every iteration. The update in $\tau_i$ iteratively changes $P$ defined in \eqref{eq:P}. The loss function and the training methods are introduced in the next section.

%%%%%%%%%%%%%%%%%%%%%%%%%%%%%%%%%%%%%%%%%%%%%%%%%%%%%%%%%%%%
\section{Training the networks with delays}\label{sec:train}
%%%%%%%%%%%%%%%%%%%%%%%%%%%%%%%%%%%%%%%%%%%%%%%%%%%%%%%%%%%%

After constructing of the NODE (\ref{eq:NODE},\ref{eq:NODE2}), one can simulate the resulting differential equations with given initial condition $X(0)$. The corresponding solution is denoted by $\widehat{X}(t)$ and we have ${\hat{x}(t) = \widehat{X}_1(t)}$. During the learning process this is compared to the data, which is obtained as the solution $x(t)$  of \eqref{eq:DDE}. We assume that the data is available with the sampling time $h$ and we compare the solutions along the time horizon ${T = N h, N \in \mathbb{N}}$. We construct the loss function using the one-norm:
\begin{equation}\label{eq:loss}
    L = \sum_{j = 1}^{N} \big\|\hat{x}(jh)-x(jh)\big\|_1,
\end{equation}
which depends on the initial history $X(0)$ and the trainable parameters ${\theta = \{W_l,b_l,\tau_i\}, l=1,\ldots,L, i = 1,\ldots,d}$. 

Learning algorithms require the gradient
\begin{equation}\label{eq:grad}
g = \frac{\partial L}{\partial \theta},    
\end{equation}
to iterate the parameters so that \eqref{eq:loss} decreases.
The gradient of the loss with respect to delay $\tau_i$ is given by
\begin{equation}
    \frac{\partial L}{\partial \tau_i} = \frac{\partial L}{\partial x(-\tau_i)} \frac{\partial x(-\tau_i)}{\partial \tau_i} = \frac{\partial L}{\partial x(-\tau_i)} \big(-\dot{x}(-\tau_i)\big),
\end{equation}
where $\frac{\partial L}{\partial x(-\tau_i)}$ is the gradient with respect to the initial delayed state and $\dot{x}(-\tau_i)$ is the time derivative of $x$ at $-\tau_i$. The latter one is approximated from data again using central differences while the former one can be calculated via back propagation or through the adjoint variable method as in~\cite{chen2018neural} after the forward simulation pass is established. 
The gradient of the loss with respect to other parameters $\frac{\partial L}{\partial W_l}$ and $\frac{\partial L}{\partial b_l}$ can be also obtained via back propagation.

With the availability of the gradient information, one may update the parameters using adaptive moment estimation (Adam,~\cite{kingma2014adam}). 
The formula for updating each individual parameter is given by
\begin{equation}\label{eq:adam1}
    \theta_{k+1} = \theta_k-\frac{\eta}{\sqrt{\hat{v}_k}+\epsilon}\hat{m}_k,
\end{equation}
with 
\begin{equation}\label{eq:adam2}
\begin{split}
    \hat{m}_k &= \frac{m_k}{1-\beta_1^k},
    \\
    \hat{v}_k &= \frac{v_k}{1-\beta_2^k},
\end{split}
\end{equation}
where $m_k$ and $v_k$ are the first moment estimate and second moment estimate, respectively.
The updates for $m_k$ and $v_k$ are given by
\begin{equation}\label{eq:adam3}
\begin{split}
    m_k &= \beta_1 m_{k-1} + (1-\beta_1) g_{k},
    \\
    v_k &=\beta_2 v_{k-1} + (1-\beta_2) g_{k}^2,
\end{split}
\end{equation}
using the gradient information $g_{k}$ at iteration $k$; cf.~\eqref{eq:grad}. The parameters $\eta,\epsilon,\beta_1,\beta_2$ in (\ref{eq:adam1},\ref{eq:adam2},\ref{eq:adam3}) can be tuned by the user to achieve better performance; see~\cite{kingma2014adam}.

The loss function described in \eqref{eq:loss} is based on one simulation with a given initial history. When using multiple trajectories (from multiple initial histories) for one update, the sum of simulation losses for all the trajectories in the batch is considered. Then the gradient is calculated by taking the average of the gradients in the batch. We remark that when updating the delay parameters, we limit the delay value to be between 0 and $\tau_{\rm max}$ using ${\tau_i = \max \big( \min ( \tau_i,\tau_{\rm max} ) ,0\big)}$. Since the delay values are explicitly learned, one may in fact use the first row of the NODE (\ref{eq:NODE},\ref{eq:NODE2}) as a neural delay differential equation (NDDE) whose form is similar to \eqref{eq:DDE}.

%%%%%%%%%%%%%%%%%%%%%%%%%%%%%%%%%%%%%%%%%%%%%%%%%%%%%%%%%%%%
\section{Mackey-Glass Equation}\label{sec:MG}
%%%%%%%%%%%%%%%%%%%%%%%%%%%%%%%%%%%%%%%%%%%%%%%%%%%%%%%%%%%%

\begin{figure}
\begin{center}
\includegraphics[width=8cm]{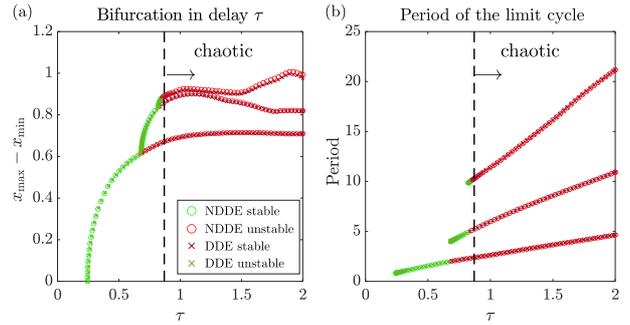}
\caption{Bifurcations diagram of the learned NDDE and the Mackey-Glass DDE. The equilibrium ${x(t)\equiv1}$ loses its stability via Hopf bifurcation at ${\tau = 0.24}$ and the arising stable periodic orbit undergoes a period-doubling cascade (with the first two bifurcations located at ${\tau = 0.61}$ and ${\tau = 0.84}$. The chaotic behavior starts at ${\tau = 0.87}$.} 
\label{fig:Bif}
\end{center}
\end{figure}

In this section, we examine the ability of the tools developed above while learning the dynamics of the Mackey-Glass equation~(\cite{mackey1977oscillation}), a scalar autonomous time delay system of the form
\begin{equation}\label{eq:MG_DDE}
\dot{x}(t) = \frac{\beta x(t-\tau)}{1+(x(t-\tau))^\delta}-\gamma x(t), 
\end{equation}
with ${\beta = 4, \gamma = 2, \delta = 9.65}$. The system exhibits different qualitative behaviors as the time delay $\tau$ is varied. For small values of $\tau$, there exists a globally stable equilibrium. This equilibrium looses stability as $\tau$ is increased via a supercritical Hopf bifurcation, giving rise to stable limit cycle oscillations. With $\tau$ increased further, the limit cycle loses stability via a supercritical period doubling bifurcation, and a period doubling cascade eventually leads to chaotic behavior; see the bifurcation diagram in Fig.~\ref{fig:Bif}(a). We set ${\tau = 1}$, which yields chaotic behavior, and generate 100 trajectories using constant initial histories ${x(t) \equiv}\ c,\ t\in[-\tau_{\rm max}, 0]$ where ${c =  0.5 + i / 99, i=  0,\ldots 99}$ and ${\tau_{\rm max}=1.5}$. The simulations are done in MATLAB using dde23.

According to (\ref{eq:AgmODE},\ref{eq:AgmODE2}) one may also generate data using the approximate ODE
\begin{equation}\label{eq:MG_ODE}
    \dot{X}(t) =
    \begin{bmatrix}  \frac{\beta X_{r}(t)}{1+\big(X_{r}(t)\big)^\delta}-\gamma X_1(t)
    \\
    D_M X(t) 
    \end{bmatrix}.
\end{equation}
This may serve as the baseline for the performance of the NODE explained below for which we use $h=0.05$ yielding ${M = \tau_{\rm max}/h =30 }$ and ${r = \tau/h + 1 =21 }$ and  matrix \eqref{eq:DM2}.

When constructing the NODE (\ref{eq:NODE},\ref{eq:NODE2}), we use two-hidden-layer trainable time delay neural networks with 5 neurons in each hidden layer and choose $\tanh(\cdot)$ as the nonlinear activation function. Thus, following \eqref{eq:network}, we can write
\begin{equation}\label{eq:MG_NODE}
    \dot{X}(t) =
    \begin{bmatrix}  W_3 \tanh{\Big(W_2 \tanh{\big(W_1 P X(t)+b_1\big)+b_2\Big)}}
    \\
    D_M X(t) 
    \end{bmatrix}.
\end{equation}
where ${W_1 \in \mathbb{R}^{5\times 2}}$, ${W_2 \in \mathbb{R}^{5\times 5}}$, ${W_3 \in \mathbb{R}^{1\times 5}}$, ${b_1, b_2 \in \mathbb{R}^{5\times 1}}$ and the $\tanh$ is applied element-wise. The matrix $P$ is chosen to be the same as in \eqref{eq:Pex}.
We focus on analyzing the case with knowledge of the number of delays. Recall that ${\tau_1=0}$ and only the second delay ${\tau_2>0}$ is learned. 
Both \eqref{eq:MG_ODE} and \eqref{eq:MG_NODE} are simulated using ode45 in MATLAB.
Finally, we extract the NDDE 
\begin{equation}\label{eq:MG_NDDE}
    \dot{x}(t) =
  W_3 \tanh{\Big(W_2 \tanh{\big(W_1 
  \begin{bmatrix}
  x(t)\\  x(t-\tau_2)
  \end{bmatrix}
  +b_1\big)+b_2\Big)}},
\end{equation}
utilizing the first row of \eqref{eq:MG_NODE} and simulate the corresponding equations using dde23.

\begin{figure}
\begin{center}
\includegraphics[width=7.5cm]{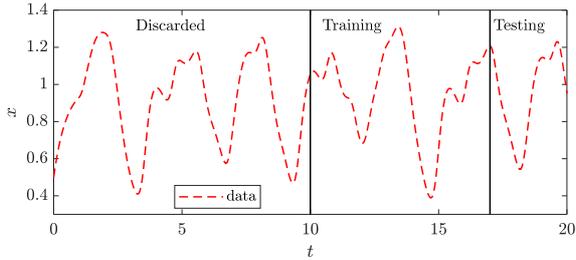}
\caption{Data obtained for training and testing. One of the 100 trajectories is shown with initial history $c = 0.5$.} 
\label{fig:DataProcess}
\end{center}
\end{figure}

When generating data via \eqref{eq:MG_DDE} we drop the transients, that is, we only consider the trajectories in the time domain ${t\in [10,20]}$ as indicated in Fig.~\ref{fig:DataProcess}. Then we sample the data with time step ${h = 0.05}$ and use the first 141 samples (${t\in [10,17]}$) for training and the next 60 samples (${t\in (17,20]}$) for testing. When calculating the loss \eqref{eq:loss}, we set the simulation horizon to be ${N = 10}$ steps (i.e., ${T = Nh = 0.5}$).
From the 141 samples in each trajectory, we take segments of 31 samples as initial histories and obtain predictions for next 10 steps. Thus, 101 input-output pairs (i.e., ${[x(t-\tau_{\rm max}),\ldots,x(t)]\rightarrow [x(t+h),\ldots,x(t+T)]}$, $t = 11.50,11.55,\ldots,17$) are generated.
This way, we obtain 10100 pairs from the 100 trajectories for training.

\begin{figure}
\begin{center}
\includegraphics[width=7.5cm]{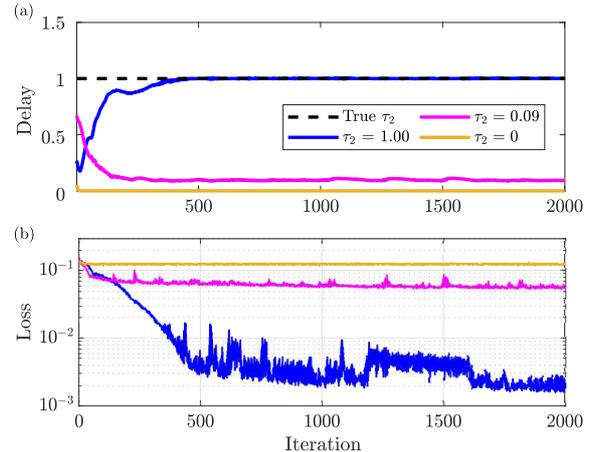}
\caption{Delay values (top) and training loss (bottom) for three different parameter initializations. The final learned delays are ${\tau_2 \approx 0, 0.09, 1}$, respectively. Note that the network parameters may jump over a local minimum and converge to another one later due to the constant learning rate, which may cause the temporary increase in training loss.}
\label{fig:Tau_Loss_comb}
\end{center}
\end{figure}

We train the network for 2000 iterations and use the batch size 1000, such that 10 out of the 101 pairs from each trajectory are used for each update. The delay $\tau_2$ is initialized uniformly between ${[0,\tau_{\rm max}]}$, $b_1$ and $b_2$ are initialized as zeros, while $W_1, W_1$ and $W_3$ are initialized using Glorot initialization~(\cite{glorot2010understanding}). 
In the Adam updating rule (\ref{eq:adam1},\ref{eq:adam2},\ref{eq:adam3}) the learning rate $\eta$ is set to 0.01 while $\beta_1,\beta_2,\epsilon$ are set to $0.9,0.999, 10^{-8}$, respectively; see~\cite{kingma2014adam}. From different parameter initializations, the networks may converge to different local minima.
We plot the delay learning path and training loss of the networks with three different learned delays as function of the iteration number in Fig.~\ref{fig:Tau_Loss_comb}. For the two paths leading to small delay values, the training loss stops decreasing very early, while for the third path the loss keeps decreasing with the iterations. This may be used to detect undesired local minima and to filter out the bad initialization at the early stages of training.

\begin{figure}
\begin{center}
\includegraphics[width=8cm]{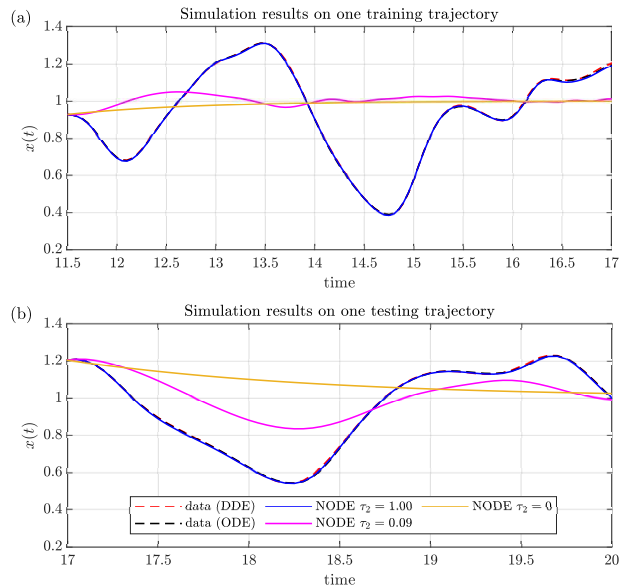}
\caption{Simulation of three trained NODEs with different learned delays for training (top) and testing (bottom).} 
\label{fig:Simu_results}
\end{center}
\end{figure}

The predictions of three trained networks are shown in Fig.~\ref{fig:Simu_results}. The simulation of the network with correctly learned delay (blue solid curve) fits the data (red and black dashed curves) very well and makes good short-term predictions about the future. The NODEs with incorrect delays (magenta and orange curves) perform poorly both for training and for testing.

\begin{figure}
\begin{center}
\includegraphics[width=8.4cm]{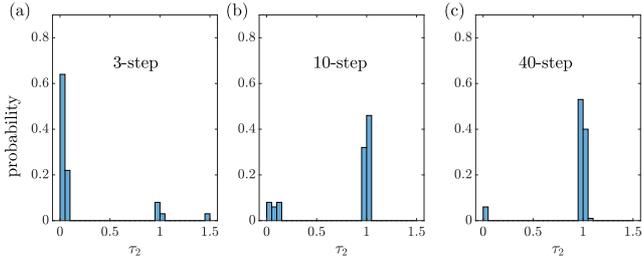}
\caption{Histograms of the learned delay at iteration 2000 when using different prediction horizons as indicated. } 
\label{fig:histo100runs_comb}
\end{center}
\end{figure}

We also study the effect of the simulation horizon on the training process. We trained 100 networks with different initial parameters using 3-step, 10-step and 40-step horizons and plot the distribution of the learned delays at iteration 2000 in Fig.~\ref{fig:histo100runs_comb}. We observe a trade-off between the performance and training time. Increasing the horizon can reduce the chance of being trapped at small delay values, but the training process is slower for longer horizons.

\begin{figure}
\begin{center}
\includegraphics[width=8.4cm]{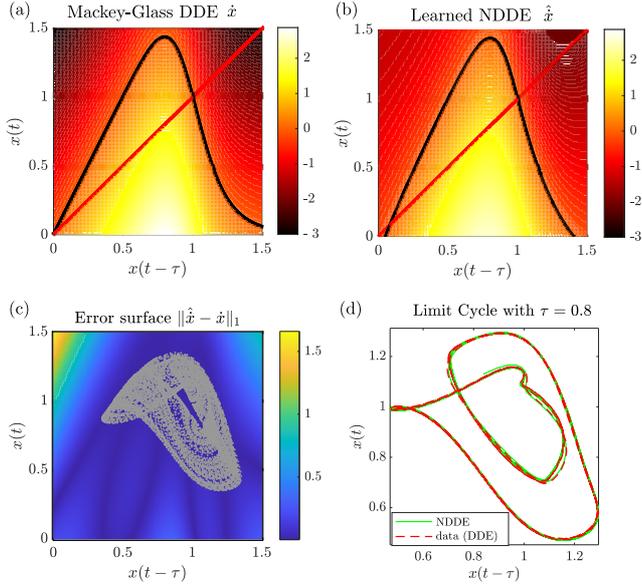}
\caption{Comparing the nonlinearity of the Mackey-Glass DDE to that of the learned NDDE (extracted from the NODE for ${\tau_2 = 1.00}$ in Fig.~\ref{fig:Simu_results}). (a), (b): top view of the nonlinear surfaces; (c): top view of the error surface, with training data indicated by gray dots; (d): limit cycle predicted by the NDDE (green curve) compared with that of DDE (dashed red curve) for ${\tau_1 = 0, \tau_2 = 0.8}$.} 
\label{fig:Nonlinear}
\end{center}
\end{figure}

To illustrate on how well the learned network performs, we compare the nonlinearity obtained through NDDE \eqref{eq:MG_NDDE} with that of the original DDE \eqref{eq:MG_DDE}. The coloring in Fig.~\ref{fig:Nonlinear}(a) shows $\dot{x}(t)$ (i.e., the right hand side of the DDE) as a function of $x(t)$ and ${x(t-\tau)}$. The black curve indicates ${\dot{x} = 0}$ and its intersection with the 45 degree red line gives the equilibria of the system, namely, ${x(t)\equiv0}$ and ${x(t)\equiv1}$. Panel (b) shows the network output $\hat{\dot{x}}(t)$ (i.e., the right hand side of the NDDE) as a function of $x(t)$ and ${x(t-\tau)}$,
while panel (c) depicts the difference between the nonlinearities of the DDE and the NDDE. The gray trajectory indicates the chaotic attractor of the DDE which covers a significant portion of the plane.

Since the time delay and the nonlinearity are learned separately, we can vary the time delay while keeping the  network with same weights and biases and analyze the behavior of the corresponding NDDE. Fig.~\ref{fig:Nonlinear}(d) depicts the trajectory for ${\tau_2 = 0.8}$ where the system approaches a stable period two limit cycle. (Recall that the training was performed for ${\tau_2 = 1}$ where the system approached a chaotic attractor.) The constant initial condition ${x(t) = 0.2}$ is used which lies outside the training dataset. The neural network gives a good prediction for the limit cycle, since the nonlinearity of the learned NDDE is similar to the of the DDE. 
This demonstrates the advantage of learning the delays in addition to learning the weights and biases. 
With the explicitly learned time delays, the network generalizes well and provides insights to the system dynamics.

We exhibit the bifurcation diagram of the trained NDDE and compare it to that of the original DDE in Fig.~\ref{fig:Bif}. One may observe the formation of limit cycles and their stability changes as the delay increases from 0 to 2. The network is able to predict appearance of Hopf and period doubling bifurcations and give good predictions for the amplitude, period and stability of the limit cycles without requiring any retraining.

\begin{figure}
\begin{center}
\includegraphics[width=7.5cm]{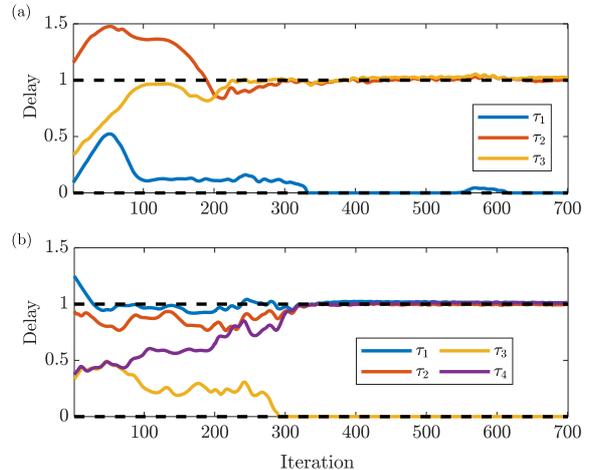}
\caption{Delay learning paths with 3 delays (top) and 4 delays (bottom). Both networks are trained with 30-step simulation horizon.} 
\label{fig:delays}
\end{center}
\end{figure}

Finally, the method still works if we do not have knowledge about the zero delay, or even about the number of the delays in the system. When (moderately) overestimating the number of delays, we can still learn the dynamics. Fig.~\ref{fig:delays} shows the delay learning paths of two different network configurations with 3 delays and with 4 delays, respectively. Observe that the delays converge to values around 0 and 1.  The nice generalization ability of the networks in this example does depend on the system and training data. Thanks to the chaotic behavior, the training data is rich enough to recover the nonlinearity locally for this autonomous single-delay scalar system. However, similar to other data-driven methods, the performance of this method is limited by the quality of the data. For autonomous systems with multiple states and/or multiple delays, obtaining good training data with rich dynamics might be challenging in many applications.

%%%%%%%%%%%%%%%%%%%%%%%%%%%%%%%%%%%%%%%%%%%%%%%%%%%%%%%%%%%%
\section{Conclusion}\label{sec:conclu}
%%%%%%%%%%%%%%%%%%%%%%%%%%%%%%%%%%%%%%%%%%%%%%%%%%%%%%%%%%%%

A method for learning the dynamics of time delay systems was presented. The key idea is to transform the dynamics described by a delay differential equation to an ordinary differential equation through discretizing the history, while retaining continuous time evolution of the systems to enable high accuracy predictions. The dynamics was learned by constructing a neural ordinary differential equation, whose right hand side, represented by a neural network with certain structure, approximated the right hand side of the ODE. A neural delay differential equation can be extracted from the obtained NODE, and it generalized very well when either the parameters or the initial conditions were changed. The developed methods were applied to the Mackey-Glass equation where both the nonlinearity and the time delay were learned with high accuracy. Future research includes applying the developed method to real engineering systems with and without first principle models.

\bibliography{TDS2022_bib}  

% \appendix
% \section{A summary of Latin grammar}    % Each appendix must have a short title.
% \section{Some Latin vocabulary}              % Sections and subsections are supported  
                                                                         % in the appendices.
\end{document}